\documentclass{sigchi-ext}
\usepackage[T1]{fontenc}
\usepackage{textcomp}
\usepackage[scaled=.92]{helvet} 
\usepackage{graphicx} 
\usepackage{balance}  
\usepackage{booktabs} 
\usepackage{ccicons}  
\usepackage{ragged2e} 

\def\plaintitle{Potential applications of Machine Learning at Multidisciplinary Medical Team Meetings} 
  
\def\emptyauthor{}
\def\plainkeywords{Machine Learning; Speech and Language Processing; Multidisciplinary Medical Team Meeting; Collaboration}

\title{Potential Applications of Machine Learning at Multidisciplinary Medical Team Meetings}

\numberofauthors{3}
\author{%
  \alignauthor{%
    \textbf{Bridget Kane}\\
 \affaddr{Karlstad University Business School}\\
 \affaddr{P.O. Box 1212}\\
  \affaddr{Karlstad, SE-651 88,} 
  \affaddr{Sweden}\\
  \email{bridget.kane@kau.se} }
  \alignauthor{%
    \textbf{Jing Su}\\
    \affaddr{School of Computer Science and Statistics}\\
    \affaddr{Trinity College Dublin}\\
  \affaddr{Dublin,}
  \affaddr{Ireland}\\
  \email{jing.su@tcd.ie}}
  \vfil \alignauthor{%
    \textbf{Saturnino Luz}\\
    \affaddr{Usher Institute}\\
    \affaddr{Edinburgh Medical School}\\
    \affaddr{The University of Edinburgh}\\
    \affaddr{Edinburgh, Scotland, UK}
    \email{s.luz@ed.ac.uk} } }

\definecolor{linkColor}{RGB}{6,125,233}
\hypersetup{%
  pdftitle={\plaintitle},
  pdfauthor={\emptyauthor},
  pdfkeywords={\plainkeywords},
  bookmarksnumbered,
  pdfstartview={FitH},
  colorlinks,
  citecolor=black,
  filecolor=black,
  linkcolor=black,
  urlcolor=linkColor,
  breaklinks=true,draft
}

\begin{document}

\CopyrightYear{2020}
\setcopyright{rightsretained}
\conferenceinfo{CSCW'19,}{November 9th-13th 2019, Austin, Texas}
\isbn{978-1-4503-6819-3/20/04}
\doi{https://doi.org/10.1145/3334480.XXXXXXX}
\copyrightinfo{\acmcopyright}

\maketitle

\RaggedRight{} 

\begin{abstract}
While machine learning (ML) systems have produced great advances in several domains, their use in support of complex cooperative work remains a research challenge. A particularly challenging setting, and one that may benefit from ML support is the work of multidisciplinary medical teams (MDTs). This paper focuses on the activities performed during the multidisciplinary medical team meeting (MDTM), reviewing their main characteristics in light of a longitudinal analysis of several MDTs in a large teaching hospital over a period of ten years and of our development of ML methods to support MDTMs, and identifying opportunities and possible pitfalls for the use of ML to support MDTMs.
\end{abstract}



\section{Introduction}

An MDT is a group of specialists from
different healthcare professions who collaborate on diagnosis and treatment of patients in their care.
An MDT for cancer, for
instance, will include physicians, surgeons, pathologists,
radiologists, medical and radiation oncologists, nurses, and other
professionals \cite{bib:KaneLuzCSCWJ06,bib:kaneluz2009adc}.  
They work independently from each other within
teams and hierarchies in their own specialist area; their interaction
is an essential part of healthcare work.

Multidisciplinary teamwork has gained importance in health over the
last decades. Since its inception in cancer care, MDT working has been
recommended for the management of other conditions such as chronic
obstructive pulmonary disease (COPD), diabetes, rheumatology
\cite{bib:VerhoefToussaintEtAl07}, and neurological
conditions \cite{bib:TaylorMunroEtAl10m}. While conclusive studies
on effectiveness of MDT care are lacking
\cite{bib:nicemethod05,bib:FleissigJenkinsEtAl06m}, there is a growing
body of observational evidence associating MDT work with
improvements in communication among specialties
\cite{bib:RuhstallerRoeEtAl06m}, decision making, patient and team
member experience, as well as medical outcomes
\cite{bib:JunorHoleGillis94m,bib:KaneLuz13dnh,bib:TaylorMunroEtAl10m}.
\marginpar{%
  \vspace{-5pt} \fbox{%
    \begin{minipage}{0.925\marginparwidth}
      \textbf{Application of AI /ML in MDT teamwork} \\
      \vspace{1pc} \textbf{Preparation:} \\
      Selection of suitable text and images from a patient's EHR. \\
      \vspace{1pc} \textbf{Materials:} \\
      Decision Support through identification of current \\
      Clinical Practice Guidelines. 
      
      \vspace{1pc} \textbf{Images \& Figures:} \\
      Finding comparison images, either from prior imaging on same patient, or from other patients with known outcomes.
      \\
      \vspace{1pc} \textbf{Recording:} \\
      Talk structures can facilitate use of ML methods to retrieve sections of discussions.
      
    \end{minipage}}\label{sec:sidebar} }
MDT work, however, is a complex, time consuming activity that causes
considerable increase in the workload of the professionals involved,
particularly those specialists who are members of several MDTs, such
as radiologists and pathologists
\cite{bib:KaneLuz07multidradiolpatholdepar}.  The economic and
organisational pressures MDTMs impose, the complexity of the teamwork involved, and the
amount of discipline-specific information exchanged by the MDT
suggest that this type of  teamwork might benefit from the support of
intelligent systems, and the variety of settings and information
exchanges would suggest the use of ML.

Having identified information and record keeping needs of MDTs in
previous work \cite{bib:KaneToussaintLuz13s,bib:LuzKane16aaai}, here we turn
our attention specifically to the role that ML might
play in enhancing teamwork, and the challenges that the introduction
of ML might bring to the organisational structure of
the MDTM and to the processes that converge in it.


\section{Methodological approach}
\label{sec:method}
The basis for our observations is a series
of ethnomethodologically-informed studies we conducted with MDTs
in a tertiary-referral teaching hospital. These studies encompassed
eight MDTs (respiratory, head and neck, urology, gynaecology,
gastro-intestinal, lymphoma, breast and dermatology) and gathered data
through observational fieldwork along the lines recommended by
\shortcite{bib:RandallHarperRouncefield07ffd}, including 28 hours of
video recordings of patient case discussions, 190 questionnaires, and
several hours of focused interviews.  Based on these data, we
identified the typical MDT workflow as comprising a number of mostly
concurrently performed activities distributed across the healthcare
environment which culminate in the MDT meeting (MDTM). The MDTM is
therefore a synchronous event in which information gathered in
pre-MDTM activities (e.g.\ radiology results) is presented by the various specialists, and from which a
number of post-MDTM tasks (i.e. implementation of MDTM
decisions)
originate \cite{bib:KaneLuzCSCWJ06}.  We focus our analysis here on information needs and the potential application for support for these MDTs with ML tools.

\section{ML support for MDT work}

The complex needs of the MDTM are not easily met by conventional information systems: the task of an MDT is demanding, and the information needs and recording requirements are complex.
We explore here the potential use of ML for MDTs. We consider ML in the context of the requirements identified in
\shortcite{bib:KaneToussaintLuz13s}. We also consider the development of automated or semi-automated analysis of medical images, specially in radiology, and its consequent impact on MDTMs \cite{bib:KaneLuz07multidradiolpatholdepar}. Further developments in digital pathology are expected to impact the work of the MDT meeting also \cite{bib:william2018review}.

\subsection{ML support presentation of relevant text information}
ML approaches to text categorisation and information extraction
have been applied in the analysis of free text in patient records, EHR, and patient safety reports, among other areas of the healthcare workflow where the MDTM is situated. These methods can be
employed in support of MDT work also, possibly in conjunction with well
developed medical ontologies to guide the text mining system.  In our
analysis of the need for an information record for shared decision
making at MDTMs \cite{bib:KaneToussaintLuz13s}, we report that, despite
growing standardisation, free textual information is likely to
continue to play an important role in documentation.  When an MDT 
adopted a structured form, consisting of multiple tabs, tables and
check-boxes, aimed at capturing essential items of information
exchanged during the MDTM for incorporation into the EHR, the
person-in-charge of entering the data eventually abandoned the
form structure and entered free-form text instead. This was due in
part to the time constraints under which the MDT operates, but
more broadly it reflects a suspicion that this could well
turn into one of those processes that Ash et Al. (2004)
\shortcite{bib:ashbergcoiera04some} describe as ``causing cognitive
overload by overemphasizing structured and `complete' information
entry or retrieval.''
    
\subsection{Support for the presentation of relevant images}

Image retrieval presents potential opportunities. Automatic image
matching techniques can be employed to retrieve similar radiology
images based on a patient's scan. They can be combined with text
mining of radiology reports to assist in case assessment. 
\shortcite{bib:NapelBeaulieuEtAl10autrcim} presents an
intelligent system that accurately retrieves CT images based on
visual similarity. Integrating such systems with MDTM work may also
improve the MDTM's educational function by providing context for post-MDTM review. 

\marginpar{%
  \vspace{-5pt} \fbox{%
    \begin{minipage}{0.925\marginparwidth}
      \textbf{Evaluation Methods} \\
      \vspace{.5pc}   {\large Has the time come to define methods for evaluation of ML systems take into account the context of use and the need to protect patient safety and privacy? \\} 
     
    \end{minipage}}\label{sec:sidebar2} }

\subsection{ML support for talk-based interaction}

It is widely acknowledged that the 
talk during an MDTM is a potentially valuable resource.  A comprehensive
record of an MDTM could be used to provide the
contextual information necessary to interpret decisions
recorded in the formal MDTM report produced. These reports are
necessarily concise. One such report might read ``{\it 36yrs. Core Rt
  breast FA 2.5cms. Path FA B2. Concordant. Reassure \& DC}''
\cite{bib:KaneToussaintLuz13s}. Reviewing a recording or transcription
of a case discussion would enable an MDT member to understand the rationale and diagnostic {\em process} that led to the formal report. Accessing
recorded unstructured meeting interaction data has been the focus of
much work in the field of {\em meeting browsing}. Systems are
proposed, for instance, that support the production of an index to
facilitate access to relevant time-based content by exploiting
natural structuring points of meeting interaction, such as writing events \cite{bib:LuzMasoodianMMM05}

\subsection{ML and support for audio content retrieval}

MDTM audio recordings contain rich information related to medical 
decision-making, and they are valuable for verification and staff 
training purposes. However, most audio recordings are set aside because 
people lack an easy way to retrieve the content of interest. Manual retrieval of case discussion from audio recordings is an onerous task.
Solutions of topic-oriented audio segmentation are proposed to automatically 
build indices on topic changes along recordings \cite{bib:SuJingKaneLuz08}. 
Thereafter, people gain access to an audio repository in a non-linear fashion 
which is highly efficient. 

To support privacy of MDTM proceedings, which is a major concern, 
we avoid using text transcriptions from MDTM recordings.
Vocalisation based acoustic and speaker 
features are introduced as innovative clues to predict topic boundaries \cite{bib:LuzSu2010}. 
We emphasise robust classification schemes with feature selection and 
achieve competitive topic segmentation accuracy \cite{bib:KaneLuzJing10c}. 
Moreover, a set of metrics is proposed to evaluate segmentation fitness
in this scheme \cite{bib:JingLuz2013}.

\subsection{ML identification of Current Clinical Practice Guidelines}
Among the MDT tasks is to identify the most appropriate and up-to-date management for a patient being discussed. Medical decisions are guided by  clinical practice guidelines and as these are being updated and modified on a regular basis, as new evidence becomes available, it is a challenge for the MDT to identify the appropriate guideline at any given time. Efforts are on-going to develop technology that can perform this task \cite{bib:FrykholmGroth11r}.

\subsection{Clinician Feedback}
Applying suitable data recording methods in conjunction with ML technologies has the potential to provide valuable feedback on trends and performance to the MDT. While decision-making can be difficult, it is currently even more difficult for the MDT to get feedback on the outcome of the decisions. Feedback on patient outcomes is a recognised objective for MDTs \cite{bib:Lamb11,bib:KaneLuzCSCWJ06}.

\section{Clinical Accountability, Patient Safety and Trust}
The unqualified adoption of ML in medical teamwork settings has implications for  clinical accountability and patient safety. Current limitations, including bias, privacy and security, and lack of transparency are a concern \cite{bib:Topol19}. Topol argues that the AI hype exceeds the state of AI science especially when it pertains to validation and readiness for implementation in patient care. He cites reports where AI output recommendations were erroneous, and potentially harmful to patients \cite{bib:Topol19}.

Applications that utilise machine-learning in healthcare, can be considered as a medical device. While on one hand assurance can be provided to clinicians to support them in their decisions, it is difficult to conduct evaluation studies to the level of the clinical trials required for new pharmaceuticals. Regulations on the adoption of ML applications in medical devices can be confusing: some argue that software for active patient monitoring are not medical devices; however, earlier lenient guidelines towards software regulation are withdrawn and unclassified software devices now require FDA approval for use in healthcare \cite{bib:FDA19}. The FDA's traditional paradigm of medical device regulation was not designed for adaptive AI and ML technologies, and the FDA mow anticipates that many of these AI and ML software changes may need a pre-market review \cite{bib:FDA19}.
Apart from anticipating improved effectiveness from the application of ML at MDTMs, it is worth exploring how the human aspect
of medicine, the doctor-patient relationship, may be affected by the increasing use of AI/ML in medical practice \cite{bib:NatureEditor19}.

\section{Conclusion}
\label{sec:conclusion}

Applying ML tools in MDTs promises to transform teamwork as we know it, and lead to potentially more effective collaboration among the medical specialists involved. However, there are indications that such tools need to be applied with caution so that clinicians and the public can feel confident that the technology is working {\it for} them. Users should have an understanding of the underlying mechanisms, and be able to recognise if errors are introduced into the systems.  We note in previous work that one of the functions of the MDTM is to improve patient safety by allowing revision of results and resolution of inconsistencies. It is tempting to imagine a perhaps not too distant future where a ML system acts as another MDT member, assisting the work of the MDT by retrieving and presenting evidence while undergoing the critical scrutiny and review that characterise the work of its human counterparts.

\balance{} 

\small 
\bibliographystyle{SIGCHI-Reference-Format}
\bibliography{ksl-cscw-ml-2019.bib}


\begin{thebibliography}{00}


\ifx \showCODEN    \undefined \def \showCODEN     #1{\unskip}     \fi
\ifx \showDOI      \undefined \def \showDOI       #1{{\tt DOI:}\penalty0{#1}\ }
  \fi
\ifx \showISBNx    \undefined \def \showISBNx     #1{\unskip}     \fi
\ifx \showISBNxiii \undefined \def \showISBNxiii  #1{\unskip}     \fi
\ifx \showISSN     \undefined \def \showISSN      #1{\unskip}     \fi
\ifx \showLCCN     \undefined \def \showLCCN      #1{\unskip}     \fi
\ifx \shownote     \undefined \def \shownote      #1{#1}          \fi
\ifx \showarticletitle \undefined \def \showarticletitle #1{#1}   \fi
\ifx \showURL      \undefined \def \showURL       #1{#1}          \fi

\bibitem{bib:ashbergcoiera04some}
{J.S. Ash}, {M. Berg}, {and} {E. Coiera}. 2004.
\newblock \showarticletitle{Some unintended consequences of information
  technology in health care: the nature of patient care information
  system-related errors}.
\newblock {\em JAMIA\/} {11}, 2 (2004), 104--112.
\newblock


\bibitem{bib:NatureEditor19}
{Editor}. 2019.
\newblock \showarticletitle{Medicine in the digital age}.
\newblock {\em Nature Medicine\/} {25}, 1 (2019), 1--1.
\newblock
\showISSN{1546-170X}


\bibitem{bib:FleissigJenkinsEtAl06m}
{A Fleissig}, {V. Jenkins}, {S. Catt}, {and} {L. Fallowfield}. 2006.
\newblock \showarticletitle{Multidisciplinary teams in cancer care: are they
  effective in the {UK}?}
\newblock {\em The Lancet Oncology\/}  {7} (2006).
\newblock
\showISSN{1470-2045}


\bibitem{bib:FDA19}
{U.S. Food} {and} {Drug Administration}. 2019.
\newblock Artificial Intelligence and Machine Learning in Software as a Medical
  Device.
\newblock   (04/02/2019 2019).
\newblock


\bibitem{bib:FrykholmGroth11r}
{O. Frykholm} {and} {K. Groth}. 2011.
\newblock \showarticletitle{References to personal experiences and scientific
  evidence during medical multi-disciplinary team meetings}.
\newblock {\em Behaviour \& Information Technology\/} {30}, 4 (2011), 455--466.
\newblock


\bibitem{bib:JunorHoleGillis94m}
{E.~J. Junor}, {D.~J. Hole}, {and} {C.~R. Gillis}. 1994.
\newblock \showarticletitle{Management of ovarian cancer: referral to a
  multidisciplinary team matters}.
\newblock {\em British Journal of Cancer\/} {70}, 2 (Aug. 1994), 363--370.
\newblock
\showISSN{0007-0920}


\bibitem{bib:KaneLuzCSCWJ06}
{B. Kane} {and} {S. Luz}. 2006.
\newblock \showarticletitle{Multidisciplinary medical team meetings: An
  analysis of collaborative working with special attention to timing and
  teleconferencing}.
\newblock {\em Computer Supported Cooperative Work\/} {15}, 5 (2006).
\newblock


\bibitem{bib:kaneluz2009adc}
{B. Kane} {and} {S. Luz}. 2009.
\newblock \showarticletitle{Achieving Diagnosis by Consensus}.
\newblock {\em Computer Supported Cooperative Work\/} {18}, 4 (2009), 357--391.
\newblock


\bibitem{bib:KaneLuz13dnh}
{B. Kane} {and} {S. Luz}. 2013.
\newblock \showarticletitle{{``Do} no harm'': Fortifying {MDT} collaboration in
  changing technological times}.
\newblock {\em Int. Journal of Medical Informatics\/} {82}, 7 (2013), 613--625.
\newblock
\showISSN{1386-5056}


\bibitem{bib:KaneLuz07multidradiolpatholdepar}
{B. Kane}, {S. Luz}, {D~Sean O'Briain}, {and} {Ronan McDermott}. 2007.
\newblock \showarticletitle{Multidisciplinary team meetings and their impact on
  work-flow in Radiology and Pathology Departments}.
\newblock {\em {BMC Medicine}\/}  {5}, Article 1741-7015 (2007), 15 pages.
\newblock
Issue 1.


\bibitem{bib:KaneLuzJing10c}
{B. Kane}, {S. Luz}, {and} {J. Su}. 2010.
\newblock \showarticletitle{Capturing multimodal interaction at medical
  meetings in a hospital setting: Opportunities and Challenges}. In {\em Procs.
  of LREC}.
\newblock


\bibitem{bib:KaneToussaintLuz13s}
{B.~T. Kane}, {P.~J. Toussaint}, {and} {S. Luz}. 2013.
\newblock \showarticletitle{Shared decision making needs a communication
  record}. In {\em Procs. of the Conf. on Computer supported cooperative work}
  {\em (CSCW '13)}. ACM, 79--90.
\newblock


\bibitem{bib:Lamb11}
{B.W. Lamb} {and} {others}. 2011.
\newblock \showarticletitle{Teamwork and team performance in multidisciplinary
  cancer teams: development and evaluation of an observational assessment
  tool}.
\newblock {\em BMJ Quality \& Safety\/}  {20} (2011), 849--856.
\newblock


\bibitem{bib:LuzKane16aaai}
{S. Luz} {and} {B. Kane}. 2016.
\newblock \showarticletitle{Perspectives on Intelligent Systems Support for
  Multidisciplinary Medical Teams}. In {\em AAAI Spring Symposium Series}.
  272--275.
\newblock


\bibitem{bib:LuzMasoodianMMM05}
{S. Luz} {and} {M. Masoodian}. 2005.
\newblock \showarticletitle{A Model for Meeting Content Storage and Retrieval}.
  In {\em 11th International Conference on Multi-Media Modeling}. IEEE,
  392--398.
\newblock


\bibitem{bib:LuzSu2010}
{S. Luz} {and} {Jing Su}. 2010.
\newblock \showarticletitle{Assessing the Effectiveness of Conversational
  Features for Dialogue Segmentation in Medical Team Meetings and in the AMI
  Corpus}. In {\em Procs. of SIGDIAL '10}. 332--339.
\newblock
\showISBNx{978-1-932432-85-5}


\bibitem{bib:NapelBeaulieuEtAl10autrcim}
{S.~A. Napel} {and} {others}. 2010.
\newblock \showarticletitle{Automated {Retrieval} of {CT} {Images} of {Liver}
  {Lesions} on the {Basis} of {Image} {Similarity}: {Method} and {Preliminary}
  {Results}}.
\newblock {\em Radiology\/} {256}, 1 (2010).
\newblock
\showISSN{0033-8419}


\bibitem{bib:nicemethod05}
{{National Institute for Health and Care Excellence (NICE)}}. 2005.
\newblock Diagnosis and treatment of lung cancer.
\newblock London.
\newblock


\bibitem{bib:RandallHarperRouncefield07ffd}
{D. Randall}, {R. Harper}, {and} {M. Rouncefield}. 2007.
\newblock {\em Fieldwork for design: theory and practice}.
\newblock Springer-Verlag.
\newblock


\bibitem{bib:RuhstallerRoeEtAl06m}
{T. Ruhstaller} {and} {others}. 2006.
\newblock \showarticletitle{The Multidisciplinary meeting: An indispensable aid
  to communication between different specialities}.
\newblock {\em European J. of Cancer\/}  {42} (2006), 2459--2462.
\newblock


\bibitem{bib:SuJingKaneLuz08}
{J. Su}, {B. Kane}, {and} {S. Luz}. 2008.
\newblock \showarticletitle{Automatic Content Segmentation of audio recordings
  at multidisciplinary medical team meetings}. In {\em Int. Conf. on Inf.
  Tech.} 309--312.
\newblock
\showISBNx{978-1-4244-2244-9}


\bibitem{bib:JingLuz2013}
{Jing {Su}} {and} {S. {Luz}}. 2013.
\newblock \showarticletitle{Can time dependencies and ensemble classification
  improve content-free dialogue segmentation?}. In {\em IEEE CogInfoCom}.
  183--188.
\newblock


\bibitem{bib:TaylorMunroEtAl10m}
{C. Taylor}, {A. Munro}, {R. Glynne-Jones}, {C. Griffith}, {and} {others}.
  2010.
\newblock \showarticletitle{Multidisciplinary team working in cancer: what is
  the evidence?}
\newblock {\em BMJ\/}  {340} (2010), c951.
\newblock


\bibitem{bib:Topol19}
{Eric~J. Topol}. 2019.
\newblock \showarticletitle{High-performance medicine: the convergence of human
  and artificial intelligence}.
\newblock {\em Nature Medicine\/} {25}, 1 (2019), 44--56.
\newblock
\showISSN{1546-170X}


\bibitem{bib:VerhoefToussaintEtAl07}
{J. Verhoef}, {Toussaint}, {and} {others}. 2007.
\newblock \showarticletitle{The impact of the implementation of a
  rehabilitation tool on the contents of the communication during
  multidisciplinary team conferences in rheumatology}.
\newblock {\em Int. J. of Medical Informatics\/} (2007), 856--863.
\newblock


\bibitem{bib:william2018review}
{W. William}, {A. Ware}, {A. Basaza-Ejiri}, {and} {J. Obungoloch}. 2018.
\newblock \showarticletitle{A review of image analysis and machine learning
  techniques for automated cervical cancer screening from pap-smear images}.
\newblock {\em Computer methods and programs in biomedicine\/}  {164} (2018).
\newblock


\end{thebibliography}

\end{document}